\documentclass{article}
\usepackage{ijcai2026}

\usepackage{times}
\usepackage{soul}
\usepackage{array}
\usepackage{siunitx}
\usepackage{url}
\usepackage[hidelinks]{hyperref}
\usepackage[utf8]{inputenc}
\usepackage[small]{caption}
\usepackage{graphicx}
\usepackage{amsmath}
\usepackage{amsthm}
\usepackage{amsfonts} 
\usepackage{multirow}
\usepackage{booktabs}
\usepackage{tcolorbox}
\usepackage{algorithm}
\usepackage{algorithmic}
\usepackage[table]{xcolor}
\usepackage{colortbl}
\usepackage{subcaption}
\usepackage{enumitem}

\usepackage[switch]{lineno}
\newtcolorbox{promptbox}[1]{
  colback=black!5!white,
  colframe=black!75!black,
  title=#1,
  fonttitle=\bfseries
}

\usepackage{listings}
\lstdefinestyle{prompt}{
  basicstyle=\footnotesize\ttfamily,
  frame=single,
  breaklines=true,
  columns=fullflexible,
  xleftmargin=0.3em,
  xrightmargin=0.3em
}

\title{Beyond Prefixes: Graph-as-Memory Cross-Attention for Knowledge Graph Completion with Large Language Models}

\author{
Ruitong Liu$^{1}$, 
Boxu Lin$^{2}$, 
Peize Li$^{3}$,
Siyuan Li$^{4,\text{\textasteriskcentered}}$, 
Yunjia Wu$^{2}$, 
Te Sun$^{5}$,
Chaohan Wu$^{2}$ \\[1.0ex]
\normalfont
$^{1}$ Peking University \quad
$^{2}$ Dalian University of Technology \quad
$^{3}$ King's College London \\
$^{4}$ Peng Cheng Laboratory \quad
$^{5}$ Shanghai Jiao Tong University \\[0.8ex]
\normalfont
ruitong.jerry@gmail.com, yuanlsy@mail.dlut.edu.cn \\[1.0ex]
\normalfont
\textasteriskcentered\ Corresponding author
}

\begin{document}
\maketitle

\begin{abstract}

Fusing Knowledge Graphs with Large Language Models (LLMs) is crucial for knowledge-intensive tasks like knowledge graph completion. Existing LLM-based approaches typically inject graph information via prefix concatenation, resulting in shallow interactions that fail to support fine-grained evidence retrieval during generation. Beyond prefixes, we propose Graph-as-Memory Tuning (GMT), a new paradigm that represents local graph structure as explicit graph memory and injects it into LLMs via deep, token-wise cross-attention. Specifically, GMT first employs a Semantic Graph Module to encode context-aware semantics from local neighborhoods guided by knowledge-enhanced relations, and compresses them into a fixed number of graph memory tokens. A Graph-as-Memory Cross-Attention Fusion Module then integrates these tokens into multiple Transformer layers, allowing LLM hidden state to dynamically retrieve relevant graph evidence. To enable efficient adaptation, GMT applies LoRA only to the memory cross-attention while keeping the base LLM frozen. Extensive experiments show that GMT significantly outperforms prefix-tuning and other strong baselines, providing more potent signals for robust reasoning. The code is published at \url{https://github.com/tongruiliu/GMT}.
\end{abstract}

\section{Introduction}
Knowledge Graphs (KGs) organize complex facts into structured triples $(\textit{h}, \textit{r}, \textit{t})$, empowering applications ranging from recommendation systems~\cite{cui2025RS,chen2025data} to question answering~\cite{omar2023universal,tkde2}. However, their inherent incompleteness necessitates Knowledge Graph Completion (KGC) to infer missing links from existing facts~\cite{tkde1,tpami2}.

Traditionally, KGC has relied on \emph{embedding-based models}, including geometric approaches~\cite{bordes2013translating,sun2019rotate} and graph neural networks~\cite{dettmers2018convolutional,zhang2021neural}. While proficient at encoding static structural patterns, these methods often overlook the rich textual semantics inherent in entities and relations. To address this, recent research has pivoted towards \emph{LLM-based paradigms}~\cite{yao2019kg,mukdc_ijcai2024,zhang2024making}, which leverage the generative and semantic capabilities of pre-trained models. However, existing integration strategies, predominantly based on prefix-tuning, typically employ a shallow fusion approach that simply concatenates structural embeddings with textual inputs~\cite{li2024cokadapter}. This shallow interaction fails to deeply align structural signals with textual representations, imposing a heavy implicit reasoning burden on the LLM and often leading to hallucinations or context-insensitive predictions.

\begin{figure}[t]
    \centering
    \includegraphics[width=1\linewidth]{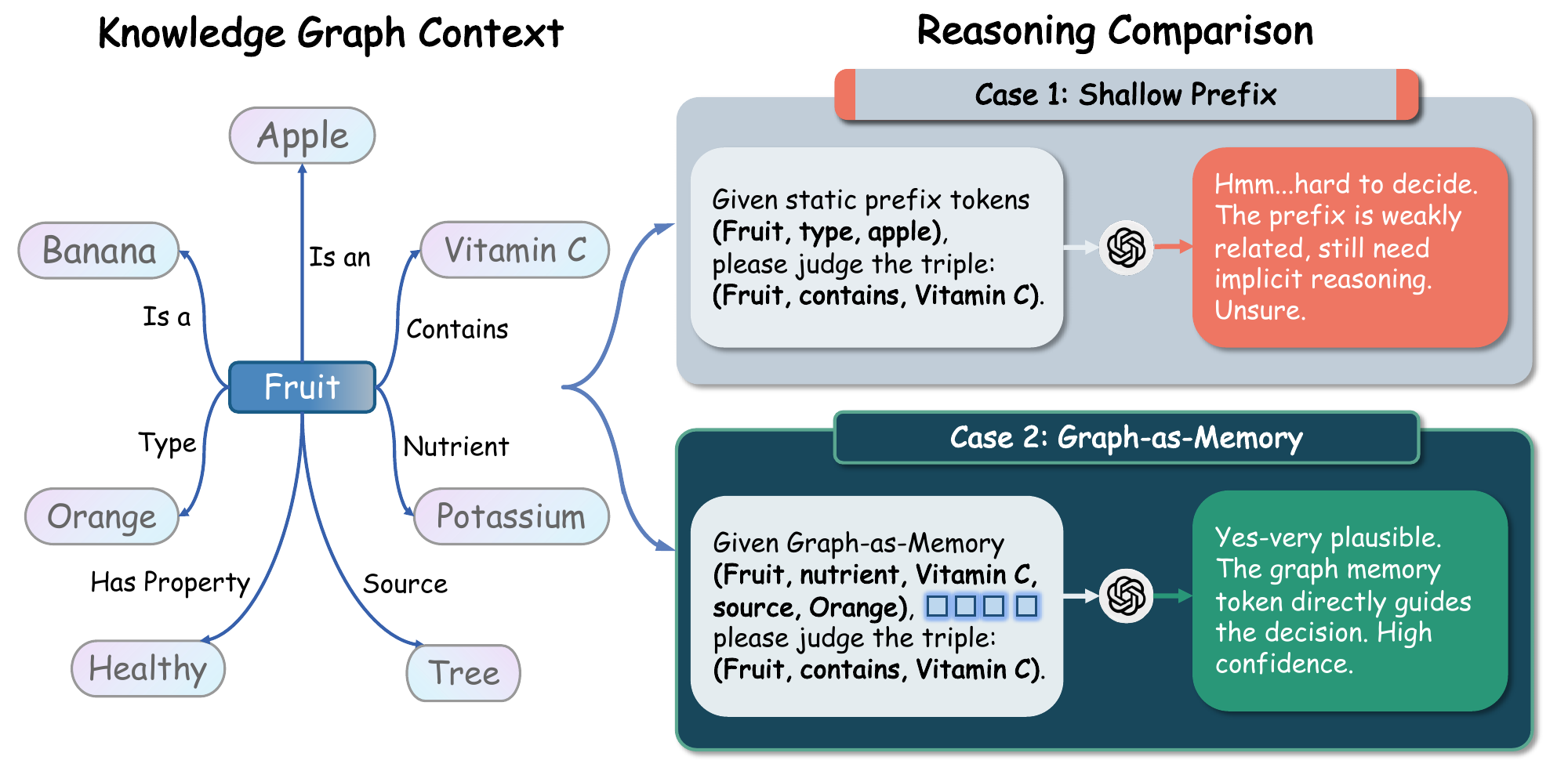}
    \caption{The guiding effect of structured information on LLMs. The semantics of the relation ``Treats'' change dynamically based on the graph context.}
    \label{fig:semantics}
    \vspace{-5mm}
\end{figure}

Consequently, a critical challenge remains: \emph{how to effectively fuse explicit KG structure with implicit LLM semantics at a deep, feature-interactive level$?$} As illustrated in Figure \ref{fig:semantics}, relational semantics are dynamic and context-dependent. The relation "Treats" implies distinct mechanisms, such as "Symptomatic Relief'' or "Pathogen Targeted Treatment", depending entirely on the local graph neighborhood (e.g., \textit{Aspirin} vs. \textit{Oseltamivir}). Capturing these nuanced shifts requires more than static concatenation; it demands a mechanism that can dynamically modulate the LLM's perception based on structural context.

To bridge this gap, we propose \textbf{Graph-as-Memory Tuning (GMT)}, a memory-centric framework that reframes local graph structure as an explicit graph memory and integrates it into LLMs through deep, token-wise cross-attention. GMT comprises two core components:
1) A \textbf{Semantic Graph Module} that uses a relation centric message passing mechanism to extract a dense and context-aware semantics from the local neighborhood, guided by knowledge enhanced relation descriptions, and compresses them into a fixed number of graph memory tokens.
2) A \textbf{Graph-as-Memory Cross-Attention Fusion Module} that injects these graph memory tokens into multiple Transformer layers, enabling each prompt token hidden representation to dynamically retrieve relevant evidence from the graph memory during generation. To maintain parameter efficiency, GMT keeps the base LLM frozen and applies LoRA only to the memory cross-attention pathway.

Extensive experiments on standard benchmarks demonstrate that GMT significantly outperforms state-of-the-art baselines while maintaining high parameter efficiency. Our main contributions are:
\begin{itemize}
    \item We propose \textbf{GMT}, a deep fusion paradigm that replaces shallow concatenation with memory-based, token-wise retrieval via cross-attention, bridging graph structure and LLM semantics. (Section \ref{sec.3})
    \item We introduce a \textbf{Semantic Graph Module} that leverages knowledge-enhanced relation semantics to guide neighborhood aggregation and construct compact graph memory tokens. (Section \ref{3.2})
    \item We design a \textbf{Graph-as-Memory Cross-Attention Fusion Module} that performs multi-layer memory injection and token-wise retrieval, and enable parameter-efficient training by applying LoRA-based adaptation to align graph memory with a frozen LLM. (Section \ref{3.3})

    \item Empirical results confirm that GMT achieves state-of-the-art performance on multiple KGC benchmarks, validating the efficacy of deep injection. (Section \ref{4})
\end{itemize}

\section{Related Work}

\subsection{Knowledge Graph Completion}
Traditional KGC methods predominantly rely on embedding-based paradigms to capture structural patterns. Geometric models, such as TransE~\cite{bordes2013translating} and RotatE~\cite{sun2019rotate}, map entities to continuous spaces using translational or rotational scoring functions. Extensions like HAKE~\cite{zhang2020learning} and BoxE~\cite{abboud2020boxe} further incorporate hierarchical and logical constraints. Parallelly, GNN-based approaches like ConvE~\cite{dettmers2018convolutional} and NBFNet~\cite{zhang2021neural} capture local topology through message passing. Despite their structural efficacy, these methods assign static representations to entities, limiting their ability to generalize to unseen data or leverage the rich semantics inherent in KGs~\cite{chang2024path}.

\subsection{LLMs for KGC}
The adoption of Large Language Models has significantly advanced semantic reasoning in KGC. Early LLM-based discriminative approaches, such as KG-BERT~\cite{yao2019kg} and SimKGC~\cite{wang2022simkgc}, formulate KGC as a sequence classification task but do not explicitly model graph topology. In the generative setting, instruction-tuned frameworks including KICGPT~\cite{kicgpt} and KG-LLaMA~\cite{yao2023exploring} directly predict tail entities, while later methods such as MKGL~\cite{guo2024mkgl} and KG-FIT~\cite{jiang2024kg} incorporate multi-view learning and few-shot inductive reasoning capabilities. More recent frameworks seek to integrate explicit structural information with the semantic reasoning capabilities of LLMs. Representative approaches include KoPA~\cite{zhang2024making} and CoK-Adapter~\cite{li2024cokadapter}, which project structural embeddings into the textual space as prefix tokens; SSQR~\cite{lin2024self}, which optimizes this via quantized representations; and GLTW~\cite{gltw2025}, which employs a joint graph transformer architecture. Nevertheless, these methods largely rely on shallow integration strategies such as prefix concatenation or textualization, failing to establish a deep, feature-level interaction in which graph structure dynamically modulates the internal representations of the LLM.

\begin{figure*}[t]
    \centering
    \includegraphics[width=0.9\linewidth]{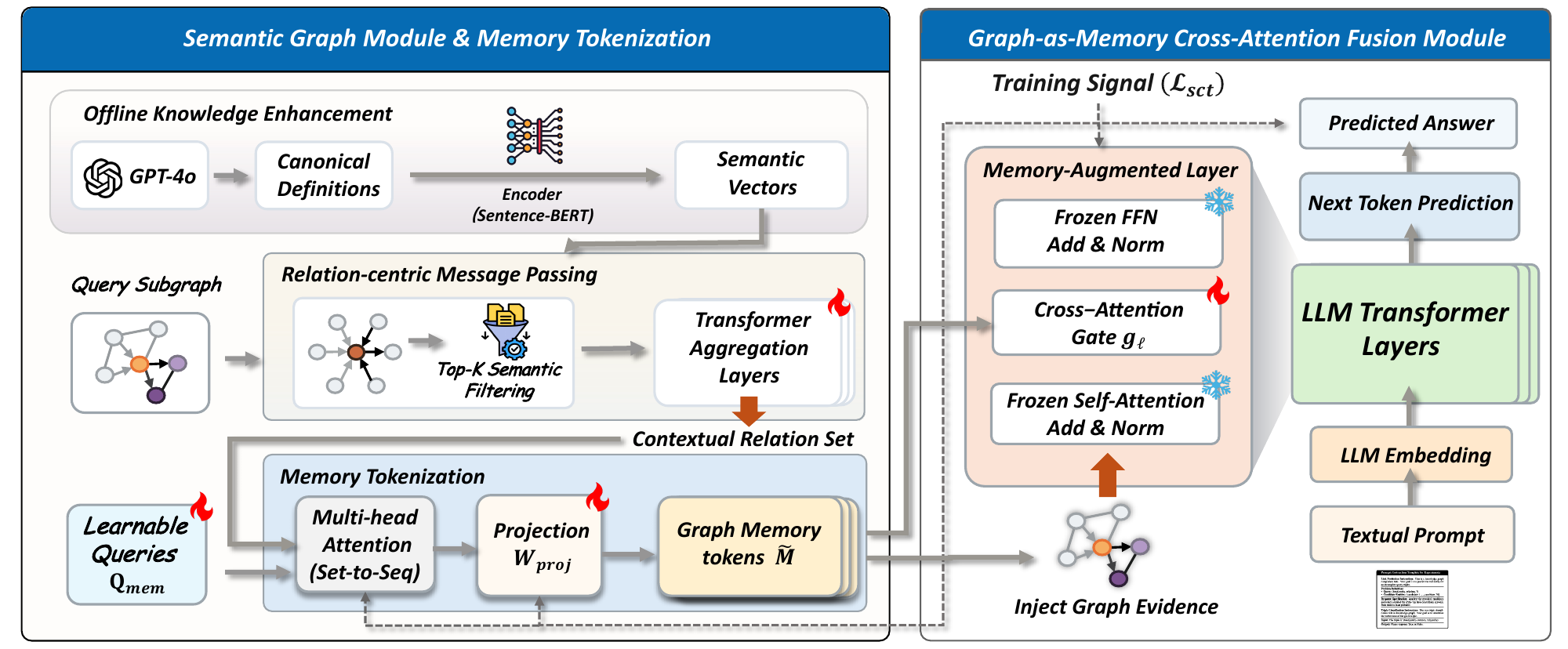}
    \caption{The GMT framework.}
    \label{fig:framework}
    \vspace{-4mm}
\end{figure*}

\section{Methodology}
\label{sec.3}

\subsection{Preliminaries}
A knowledge graph $\mathcal{G}$ is defined as a set of triples $\mathcal{T} = \{(h, r, t) \mid h, t \in \mathcal{E}, r \in \mathcal{R}\}$, where $\mathcal{E}$ and $\mathcal{R}$ denote the entity and relation sets, respectively. Knowledge Graph Completion (KGC) aims to infer a missing element in an incomplete triple, e.g., predicting $t$ for a query $(h, r, ?)$.
In LLM-based KGC, a model $\mathcal{M}$ is prompted with a textualized query and trained to generate (or select) the missing entity. 

As shown in \autoref{fig:framework}, our GMT conditions the LLM on a graph memory built from the local semantic subgraph around the query entities, and injects such memory into multiple Transformer layers through cross-attention, enabling deep and token-wise retrieval of graph evidence during generation.

\subsection{Semantic Graph Module}
\label{3.2}

The first stage of GMT is to transform the local neighborhood structure into a dense set of contextual semantic representations that can serve as external memory for the LLM. Unlike methods that directly rely on pre-trained entity embeddings~\cite{zhang2024making}, our Semantic Graph Module (SGM) performs \textit{relation-centric} message passing to extract semantically filtered relational evidence around the entities.

\paragraph{Relation-centric Message Passing.}
Inspired by~\cite{wang2021relational,li2025flow}, we treat relations as the primary carriers of KG semantics\footnote{For instance, the rule $(\texttt{A, is\_father\_of, B})\land(\texttt{C, is\_wife\_of, A})\rightarrow(\texttt{C, is\_mother\_of, B})$ is resolved by relational interplay. Moreover, an entity's identity is often implicitly encoded by its relational context, making relation-centric aggregation effective.}. For a given query triple $(h,r,t)$, we extract the local neighborhoods around $h$ and $t$ (masking the predicted entity for link prediction). For each central edge $e_c$ incident to $h$ or $t$, we aggregate information from its neighboring edges $e_n \in \mathcal{N}(e_c)$.
To mitigate noise from indiscriminate aggregation, we perform Top-$K$ neighbor filtering using explicit semantic relevance. Specifically, we first conduct an offline \textbf{Knowledge Enhancement} step for each relation type using a strong LLM (GPT-4o\footnote{\url{https://github.com/openai/openai-python}.}), producing canonical definitions (Prompt in~\autoref{prompt}), and encode them with a embedding model (e.g., Sentence-BERT\footnote{https://github.com/UKPLab/sentence-transformers}~\cite{sentencebert}) to obtain a semantic vector for each relation.
During message passing, for each central edge $e_c$ and a neighbor edge $e_n$, we compute relevance by cosine similarity between their pre-computed semantic vectors $\mathbf{s}_c$ and $\mathbf{s}_n$:
\begin{equation}
    \operatorname{Score}(e_c, e_n) = \frac{\mathbf{s}_c \cdot \mathbf{s}_n}{\left\|\mathbf{s}_c\right\|_2 \left\|\mathbf{s}_n\right\|_2}.
    \label{eq:score_cosine}
\end{equation}
We then select $\mathcal{N}_K(e_c)$ as the Top-$K$ neighbors with highest scores and aggregate their states (e.g., mean pooling) to obtain $\mathbf{\bar{s}}^{\mathcal{N}_K(e_c)}$.

We refine each central edge representation via a Transformer-style update, where $\mathbf{s}_c^l$ attends to the aggregated neighborhood context:
\begin{equation}
\mathbf{s}_c^{l+1} = \text{TransformerLayer}(\mathbf{s}_c^l, \mathbf{\bar{s}}^{\mathcal{N}_K(e_c)}),
\end{equation}
and iterate for $L$ layers to obtain contextual edge representations around $h$ and $t$.

\paragraph{Graph Memory Tokenization.}
A single pooled vector inevitably bottlenecks diverse evidence. Instead, we expose a \textbf{set of memory tokens} to the LLM. Let
$
\mathcal{S} = \{\mathbf{s}_{h,i}^L\}_i \cup \{\mathbf{s}_{t,j}^L\}_j \in \mathbb{R}^{N \times d_c}
$
denote the set of contextual relation states collected from both sides (with $N$ variable per query). We compress $\mathcal{S}$ into a fixed-length graph memory with $m$ tokens using a learnable set-to-sequence tokenizer. Concretely, we introduce $m$ learnable memory queries $\mathbf{Q}_{mem}\in\mathbb{R}^{m\times d_c}$ and compute:
\begin{equation}
\mathbf{M} = \text{Attn}(\mathbf{Q}_{mem},\, \mathcal{S},\, \mathcal{S}) \in \mathbb{R}^{m \times d_c},
\label{eq:memory_tokenizer}
\end{equation}
where $\text{Attn}(\cdot)$ is standard multi-head attention. $\mathbf{M}$ serves as a compact yet expressive \textbf{Graph-as-Memory} representation for the subsequent fusion module. Finally, we project $\mathbf{M}$ to the LLM hidden size:
\begin{equation}
\tilde{\mathbf{M}} = \mathbf{M}\mathbf{W}_{proj}, \quad \mathbf{W}_{proj}\in\mathbb{R}^{d_c\times d_{llm}}.
\end{equation}

\subsection{Graph-as-Memory Cross-Attention Fusion}
\label{3.3}

The core challenge is to inject the structured graph evidence into the LLM in a \emph{deep} and \emph{token-wise} manner. We propose a Graph-as-Memory Cross-Attention Fusion Module that allows each token to retrieve relevant KG evidence from $\tilde{\mathbf{M}}$ at multiple layers.

\paragraph{Memory-augmented Transformer Layers.}
Given the input prompt embeddings $\mathbf{X}=\{\mathbf{x}_1,\ldots,\mathbf{x}_n\}$, the LLM produces hidden states $\mathbf{H}^0=\mathbf{X}$ and updates them through $L_{llm}$ Transformer layers. For a subset of layers $\mathcal{L}_{mem}\subseteq \{1,\ldots,L_{llm}\}$, we insert a cross-attention sub-layer after self-attention:
\begin{equation}
\hat{\mathbf{H}}^\ell = \mathbf{H}^\ell + \text{SelfAttn}(\text{LN}(\mathbf{H}^\ell)),
\end{equation}
\begin{equation}
\mathbf{H}^{\ell+1} = \hat{\mathbf{H}}^\ell + g_\ell \cdot \text{CrossAttn}\bigl(\hat{\mathbf{H}}^\ell,\, \tilde{\mathbf{M}}\bigr),
\label{eq:cross_attn_fusion}
\end{equation}
followed by the standard FFN block. Here $\text{CrossAttn}(\cdot)$ uses queries from token states and keys/values from the graph memory, enabling each token to selectively retrieve graph evidence. We use a learnable gate $g_\ell$ for training stability, allowing the model to gradually incorporate memory.

\paragraph{Low-Rank Adaptation of Cross-Attention.}
To adapt the memory-reading pathway efficiently, we apply LoRA to the projection matrices of cross-attention (and only these parameters, unless otherwise stated). For each projection matrix $\mathbf{W}\in\{\mathbf{W}_q,\mathbf{W}_k,\mathbf{W}_v,\mathbf{W}_o\}$ in the cross-attention module, LoRA parameterizes:
\begin{equation}
\mathbf{W} = \mathbf{W}_0 + \Delta \mathbf{W},\quad \Delta \mathbf{W}=\mathbf{B}\mathbf{A},
\end{equation}
where $\mathbf{W}_0$ is frozen, $\mathbf{A}\in\mathbb{R}^{r\times d}$, $\mathbf{B}\in\mathbb{R}^{d\times r}$, and $r\ll d$. Only $\mathbf{A},\mathbf{B}$ are trained. This design keeps the base LLM frozen while opening a dedicated, parameter-efficient channel to align graph memory with the LLM latent space.

At inference, we construct graph memory $\tilde{\mathbf{M}}$ from the query neighborhood and generate the missing entity autoregressively:
\begin{equation}
\mathcal{A} = \arg\max_{\mathcal{A}} P_{\mathcal{M}}(\mathcal{A}\mid \mathcal{I}_{\text{GMT}},\, \mathbf{X},\, \tilde{\mathbf{M}}),
\label{eq:GMT_inference}
\end{equation}
where $\mathcal{I}_{GMT}$ is the instruction template and $\mathcal{A}$ is the predicted answer.

\subsection{Training Strategy}
\label{3.4}

We adopt a two-stage training paradigm. Stage 1 pre-trains the graph module to capture structural/semantic regularities, providing a strong initialization. Stage 2 aligns the graph memory and the LLM through memory-augmented fine-tuning with LoRA on cross-attention.

\subsubsection{Stage 1: Self-Supervised Pre-training }\label{stage1}
We first pre-train SGM on a self-supervised link prediction objective. Given $(h,r,t)$, the graph module produces contextualized representations for the head and tail, denoted as $\mathbf{e}_h$ and $\mathbf{e}_t$ (obtained by mean pooling the final relation states associated with each entity). We define a plausibility score:
\begin{equation}
F_{\mathcal{G}}(h,r,t) = \langle \mathbf{e}_h, \mathbf{e}_r, \mathbf{e}_t \rangle,
\end{equation}
where $\mathbf{e}_r$ is the knowledge-enhanced relation semantic vector (from GPT-4o and Sentence-BERT).
With negative sampling, for each positive triple $(h,r,t)\in\mathcal{T}$, we construct corrupted triples $\mathcal{T}'=\{(h'_i,r,t'_i)\}_{i=1}^k$. The loss is:
{\footnotesize
\begin{equation}
\mathcal{L}_{\text{Graph}}
= - \sum_{(h,r,t)\in\mathcal{T}}
\Bigl[
\log\sigma(F_{\mathcal{G}}(h,r,t))
+ \sum_{i=1}^{k}\log\sigma\!\bigl(-F_{\mathcal{G}}(h'_i,r,t'_i)\bigr)
\Bigr].
\end{equation}
}
This stage equips SGM with robust relational semantics before interacting with the LLM.







\subsubsection{Stage 2: Memory-Augmented Alignment with the LLM}
\label{stage2}
Starting from the pre-trained SGM, we fine-tune the full GMT pipeline on the KGC objective while keeping the base LLM frozen. For each training query, we build the graph memory $\tilde{\mathbf{M}}$ using Eq.~\eqref{eq:memory_tokenizer} and feed the textual prompt into the LLM equipped with memory cross-attention layers (Eq.~\eqref{eq:cross_attn_fusion}). We optimize the auto-regressive next-token prediction loss:
\begin{equation}
\mathcal{L}_{\text{GMT}} = - \sum_{i=1}^{|\mathcal{S}_{GMT}|} \log P(s_i \mid s_{<i}),
\end{equation}
where $\mathcal{S}_{GMT} = \mathcal{I}_{GMT} \oplus \mathbf{X} \oplus \mathcal{A}$. Gradients from $\mathcal{L}_{\text{GMT}}$ update: (i) the graph memory tokenizer parameters and projection $\mathbf{W}_{proj}$, and (ii) the LoRA weights of cross-attention in the selected layers. This training aligns the graph memory space with the LLM’s internal representations through a dedicated memory-reading pathway, enabling deep fusion without modifying the base model weights.

\begin{table}[t]
\centering
\caption{Statistics of the benchmark datasets. For triple classification datasets (UMLS, CoDeX-S, FB15k-237N), the validation/test splits are denoted as positive/negative samples.}
\label{tab:dataset_stats}
\resizebox{\columnwidth}{!}{%
\begin{tabular}{l l r r r r r}
\toprule
\textbf{Dataset} & \textbf{Task} & \textbf{\#Entities} & \textbf{\#Relations} & \textbf{\#Train} & \textbf{\#Valid} & \textbf{\#Test} \\
\midrule
WN18RR & Link & 40,943 & 11 & 86,835 & 3,034 & 3,134 \\
FB15k-237 & Link & 14,541 & 237 & 272,115 & 17,535 & 20,466 \\
\midrule
UMLS & Triple & 135 & 46 & 5216 & 652/652 & 661/661 \\
CoDeX-S & Triple & 2034 & 42 & 32888 & 1827/1827 & 1828/1828 \\
FB15k-237N & Triple & 13,104 & 93 & 87,282 & 7041/7041 & 8226/8226 \\
\bottomrule
\end{tabular}%
}
\end{table}


\section{Experiments}
\label{4}

To evaluate the effectiveness of GMT, we designed
a series of experiments to address the following research
questions:
\begin{itemize}
  \setlength{\itemsep}{0pt}
  \setlength{\parskip}{0pt}
  \setlength{\topsep}{0pt}
  \setlength{\partopsep}{0pt}
  \setlength{\parsep}{0pt}
    \item \textbf{RQ1:} How significantly can GMT enhance LLMs’ performance in KGC tasks$?$
    \item \textbf{RQ2:} What are the distinct contributions of GMT’s key components to its overall performance?
    \item \textbf{RQ3:} How Does Context from Semantic Graphs Enhance the Reasoning of LLMs?
\end{itemize}

\begin{table*}[ht]
\small
\centering
\caption{Link Prediction on FB15k-237 and WN18RR datasets. Best results are in \textbf{bold}, second best are \underline{underlined}.}
\renewcommand{\arraystretch}{0.9}
\setlength{\tabcolsep}{8pt}
\label{tab:link_prediction}
\definecolor{LightGray}{gray}{0.95}
\begin{tabular}{c|c|>{\columncolor{LightGray}}cccc|>{\columncolor{LightGray}}cccc}
\toprule
\multirow{2}{*}{\textbf{Type}} & \multirow{2}{*}{\textbf{Model}} & \multicolumn{4}{c|}{\textbf{WN18RR}} & \multicolumn{4}{c}{\textbf{FB15k-237}} \\
\cmidrule(lr){3-6} \cmidrule(lr){7-10}
& & \textbf{MRR} & \textbf{Hits@1} & \textbf{Hits@3} & \textbf{Hits@10} & \textbf{MRR} & \textbf{Hits@1} & \textbf{Hits@3} & \textbf{Hits@10} \\
\midrule
\multirow{6}{*}{\textbf{Emb-based}} & TransE      & 0.223 & 0.014 & 0.401 & 0.529 & 0.330 & 0.231 & 0.369 & 0.528 \\
& CompGCN     & 0.479 & 0.443 & 0.494 & 0.546 & 0.355 & 0.264 & 0.390 & 0.535 \\
& AdaProp     & 0.562 & 0.499 & --    & 0.671 & 0.417 & 0.331 & --    & 0.585 \\
& MA-GNN      & 0.565 & 0.507 & 0.592 & 0.679 & 0.379 & 0.282 & 0.415 & 0.569 \\
& TCRA        & 0.496 & 0.457 & 0.511 & 0.574 & 0.367 & 0.275 & 0.403 & 0.554 \\
& DiffusionE  & 0.557 & 0.504 & --    & 0.658 & 0.376 & 0.294 & --    & 0.539 \\
\midrule
\multirow{7}{*}{\textbf{LLM-based}} & KICGPT        & 0.549 & 0.474 & 0.585 & 0.641 & 0.412 & 0.327 & 0.448 & 0.554 \\
& CSProm-KG-CD  & 0.559 & 0.508 & 0.578 & 0.660 & --    & --    & --    & --    \\
& ARR           & 0.521 & --    & 0.607 & --    & 0.398 & --    & 0.436 & --    \\
& KG-FIT        & 0.553 & 0.488 & 0.595 & \underline{0.695} & 0.362 & 0.275 & 0.485 & 0.572 \\
& MKGL          & 0.552 & 0.500 & 0.577 & 0.656 & 0.415 & 0.325 & 0.454 & 0.591 \\
& SSQR-LLaMA2   & 0.591 & 0.548 & 0.618 & 0.673 & 0.449 & \underline{0.374} & \underline{0.491} & 0.597 \\
& GLTW\(_{7b}\) & \underline{0.593} & \underline{0.556} & \underline{0.649} & 0.690 & \underline{0.469} & 0.351 & 0.481 & \underline{0.614} \\
\midrule
\textbf{Ours} & \textbf{GMT} & \textbf{0.621} & \textbf{0.569} & \textbf{0.667} & \textbf{0.703} & \textbf{0.488} & \textbf{0.394} & \textbf{0.505} & \textbf{0.629} \\
\bottomrule
\end{tabular}
\end{table*}

\begin{table*}[ht]
\small
    \centering
    \caption{Triple classification on UMLS, CoDeX-S, and FB15K-237N datasets. Best results are in \textbf{bold}, second best are \underline{underlined}.}
    \label{tab:main_reasults}
    \setlength{\tabcolsep}{5pt}
    \renewcommand{\arraystretch}{0.97}
    \definecolor{LightGray}{gray}{0.95}
    \begin{tabular}{c|c|>{\columncolor{LightGray}}c c c>{\columncolor{LightGray}}c|>{\columncolor{LightGray}}c c c>{\columncolor{LightGray}}c|>{\columncolor{LightGray}}c c c>{\columncolor{LightGray}}c}
    \toprule
    \multirow{2}{*}{\textbf{Type}} & \multirow{2}{*}{\textbf{Model}} & \multicolumn{4}{c|}{\textbf{UMLS}} & \multicolumn{4}{c|}{\textbf{CoDeX-S}} & \multicolumn{4}{c}{\textbf{FB15K-237N}}\\
    \cmidrule(lr){3-6} \cmidrule(lr){7-10} \cmidrule(lr){11-14} 
        & & \textbf{Acc} & 
        \textbf{P} & 
        \textbf{R} & 
        \textbf{F1} & 
        \textbf{Acc} &  
        \textbf{P} & 
        \textbf{R} & 
        \textbf{F1} & 
        \textbf{Acc} & 
        \textbf{P} & 
        \textbf{R} & 
        \textbf{F1} \\
    \midrule
    \multicolumn{1}{c|}{\multirow{4}{*}{\centering \textbf{Emb-based}}} & TransE & 84.49 & 86.53 & 81.69 & 84.04 & 72.07 & 71.91 & 72.42 & 72.17 & 69.71 & 70.80 & 67.11 & 68.91 \\ & DistMult & 86.38 & 87.06 & 86.53 & 86.79 & 66.79 & 69.67 & 59.46 & 64.16 & 58.66 & 58.98 & 56.84 & 57.90 \\ & ComplEx & 90.77 & 89.92 & 91.83 & 90.87 & 67.64 & 67.84 & 67.06 & 67.45 & 65.70 & 66.46 & 63.38 & 64.88 \\ & RotatE & 92.05 & 90.17 & 94.41 & 92.23 & 75.68 & 75.66 & 75.71 & 75.69 & 68.46 & 69.24 & 66.41 & 67.80 \\
    \midrule
    \multicolumn{1}{c|}{\multirow{6}{*}{\centering \textbf{LLM-based}}} & $\text{Alpaca}_{zero-shot}$ & 52.64 & 51.55 & 87.69 & 64.91 & 50.62 & 50.31 & \textbf{99.83} & 66.91 & 56.06 & 53.32 & \underline{97.37} & 68.91 \\ & $\text{GPT-3.5}_{zero-shot}$ & 67.58 & 88.04 & 40.71 & 55.67 & 54.68 & 69.13 & 16.94 & 27.21 & 60.15 & 86.62 & 24.01 & 37.59 \\ & $\text{ICL}_{8-shot}$ & 55.52 & 55.85 & 52.65 & 54.21 & 50.62 & 50.31 & \textbf{99.83} & 66.91 & 59.23 & 57.23 & 73.02 & 64.17 \\
    & KG-LLaMA & 85.77 & 87.84 & 83.05 & 85.38 & 79.43 & 78.67 & 80.74 & 79.69 & 74.81 & 67.37 & 96.23 & 79.25 \\ 
    & KG-Alpaca & 86.01 & \textbf{94.91} & 76.10 & 84.46 & 80.25 & 79.38 & 81.73 & 80.54 & 69.91 & 62.71 & \textbf{98.28} & 76.56 \\
    & KG-Alpaca & 86.01 & \textbf{94.91} & 76.10 & 84.46 & 80.25 & 79.38 & 81.73 & 80.54 & 69.91 & 62.71 & \textbf{98.28} & 76.56 \\
    & KoPA & \underline{92.58} & 90.85 & \underline{94.70} & 92.70 & 82.74 & 77.91 & 91.41 & 84.11 & 77.65 & 70.81 & 94.09 & 80.81 \\
    & SAT & 92.24 & 91.05 & 93.99 & \underline{93.17} & \underline{85.55} & \textbf{83.38} & 89.31 & \underline{86.54} & \underline{82.71} & \underline{82.30} & 85.24 & \underline{83.28} \\
        \midrule
\textbf{Ours} & \textbf{GMT} & \textbf{94.55} & \underline{91.86} & \textbf{95.74} & \textbf{93.76} & \textbf{89.01} & \underline{83.27} & \underline{92.43} & \textbf{87.61} & \textbf{84.10} & \textbf{83.14} & 91.27 & \textbf{87.02} \\    
    \bottomrule
    \end{tabular}
\end{table*}

\subsection{Experiment Settings}

\paragraph{Datasets.}

We evaluate our GMT framework on various widely recognized KGC benchmark datasets:

\textbf{WN18RR}~\cite{dettmers2018convolutional} and \textbf{FB15k-237}~\cite{toutanova-chen-2015-observed}: These datasets are used for the \textbf{link prediction} task. Both have inverse relations removed, making them standards for evaluating complex reasoning abilities.

\textbf{UMLS}, \textbf{CoDeX-S} and \textbf{FB15k-237N}~\cite{lv2022do}: These are employed for the \textbf{triple classification} task. It provides high-quality negative samples for each triple, making them ideal for assessing a model's ability to distinguish factual correctness.
Detailed statistics are provided in Table~\ref{tab:dataset_stats}.

\paragraph{Baselines.}
Our selection of baselines covers a wide spectrum of representative methods, from traditional KGE models to the latest LLM-based approaches. Many results are directly cited from the SSQR~\cite{lin2024self}, Kopa~\cite{zhang2024making} and GLTW~\cite{gltw2025} to maintain consistency. 

\textbf{For Link Prediction.} We compare against (1) \textit{Emb-based Methods}: TransE~\cite{bordes2013translating}, CompGCN~\cite{vashishth2020composition}, AdaProp~\cite{zhang2024adaprop}, MA-GNN~\cite{xu2023double}, TCRA~\cite{guo2024time}, and DiffusionE~\cite{ca2024diffusione}; (2) \textit{LLM-based Methods}: KICGPT~\cite{kicgpt}, CSProm-KG-CD~\cite{li2023csprom}, ARR~\cite{chen2024arr}, KG-FIT~\cite{jiang2024kgfit}, 
MKGL~\cite{guo2024mkgl}, SSQR-LLaMA2~\cite{lin2024self}.

\textbf{For Triple Classification.} We compare against (1) \textit{Emb-based Methods}: TransE~\cite{bordes2013translating}, DistMult~\cite{yang2014embedding}, ComplexE~\cite{trouillon2016complex}, and RotatE~\cite{sun2019rotate}; (2) \textit{LLM-based Methods}: KG-LLaMA~\cite{yao2023exploring}, KG-Alpaca~\cite{yao2023exploring}, KOPA~\cite{zhang2024making}, and SSQR-LLaMA2~\cite{lin2024self}.

\paragraph{Implementation Details.}

We build GMT on Alpaca-7B with the base LLM frozen. We pre-train the SGM for 20 epochs with self-supervised link prediction using negative sampling ($k{=}64$), and then fine-tune GMT with the base LLM frozen. We use $K{=}8$ for Top-$K$ neighbor filtering and $m{=}32$ memory tokens, injecting memory via cross-attention in the top layers $\mathcal{L}_{mem}{=}\{25,\dots,32\}$ with a learnable gate initialized to zero. In Stage~2, we train only the memory tokenizer, the memory projection, and LoRA on cross-attention projections $\{W_q,W_k,W_v,W_o\}$ (rank $r{=}64$, $\alpha{=}128$, dropout 0.01). We tune epochs in \{3,4,5\} and learning rate in \{$1\mathrm{e}{-4},3\mathrm{e}{-4},5\mathrm{e}{-4}$\}, using AdamW with batch size 12, bf16, and gradient clipping 100.0. Experiments are run on Nvidia A800-80GB GPUs. All closed-source models used for knowledge enhancement are accessed via their official APIs, using default inference hyperparameters (e.g., temperature) with no additional tuning.

\subsection{Overall Performance Comparison: RQ1}

\paragraph{Link Prediction}
Classic link prediction is a ranking task that evaluates the position of the gold entity among candidates. We fine-tune GMT with an instruction template $\mathcal{I}_{GMT}$ (\autoref{prompt}) and follow SSQR-LLaMA2’s candidate setting by using a pre-trained AdaProp to retrieve 20 candidates per query. As shown in Table~\ref{tab:link_prediction}, GMT achieves state-of-the-art results on both WN18RR and FB15k-237.
On \textbf{WN18RR}, GMT achieves state-of-the-art performance with \textbf{0.621} MRR and \textbf{0.703} Hits@10, outperforming the strongest LLM-based baseline (MRR 0.593). On \textbf{FB15k-237}, GMT also ranks first with \textbf{0.488} MRR and \textbf{0.629} Hits@10, surpassing the best competitor (MRR 0.469). These consistent gains across both datasets validate the effectiveness of deep, memory-based graph evidence retrieval.

\begin{table}[t]
\small
\centering
\caption{The ablation results for the link prediction task.}
\label{tab:ablation}
    \renewcommand{\arraystretch}{1.0}
    \setlength{\tabcolsep}{7pt}
\definecolor{lightgray}{gray}{0.95}
\begin{tabular}{c|cccc}
\hline
\textbf{Model} & \textbf{MRR} & \textbf{Hits@1} & \textbf{Hits@3} & \textbf{Hits@10} \\ \hline
\rowcolor{lightgray}
\multicolumn{5}{c}{\textbf{WN18RR}} \\ \hline
GMT & \textbf{0.621} & \textbf{0.569} & \textbf{0.667} & \textbf{0.703} \\
w/o Semantics & 0.603 & 0.541 & 0.639 & 0.695 \\
w/o Fusion & 0.558 & 0.516 & 0.607 & 0.639 \\ \hline\hline
\rowcolor{lightgray}
\multicolumn{5}{c}{\textbf{FB15k-237}} \\ \hline
GMT & \textbf{0.488} & \textbf{0.394} & \textbf{0.505} & \textbf{0.629} \\
w/o Semantics & 0.464 & 0.368 & 0.479 & 0.602 \\
w/o Fusion & 0.425 & 0.331 & 0.450 & 0.584 \\ \hline
\end{tabular}
\end{table}

\begin{figure*}[t]
    \centering
    \small
    \setlength{\tabcolsep}{0pt}
    \begin{subfigure}[b]{0.33\textwidth}
        \centering
        \includegraphics[width=\textwidth]{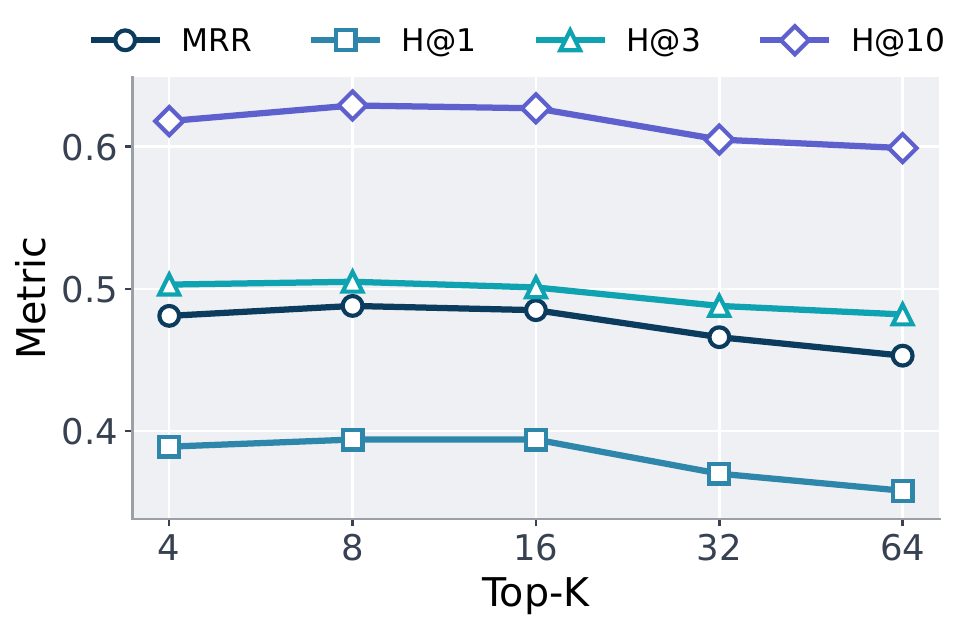}
    \end{subfigure}%
    \hfill
    \begin{subfigure}[b]{0.33\textwidth}
        \centering
        \includegraphics[width=\textwidth]{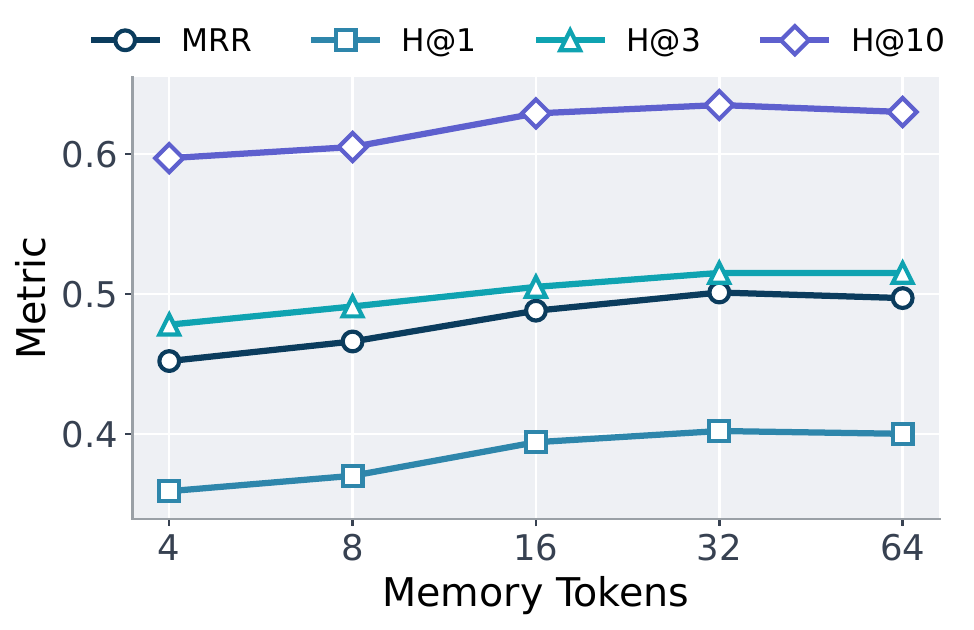}
    \end{subfigure}%
    \hfill
    \begin{subfigure}[b]{0.33\textwidth}
        \centering
        \includegraphics[width=\textwidth]{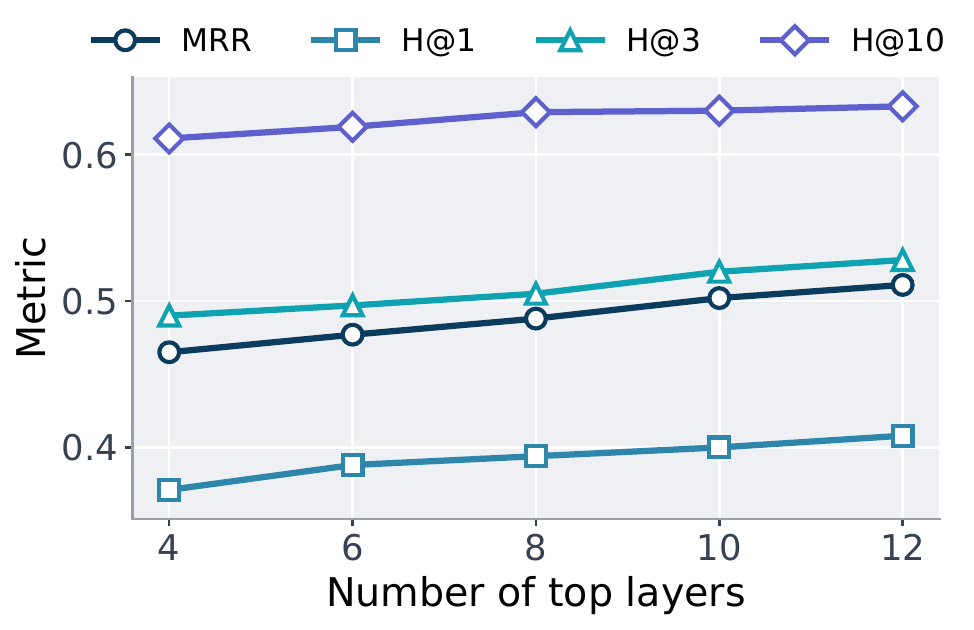} 
    \end{subfigure}
    \caption{Impact of Key Hyperparameters on GMT Performance}
    \label{fig:hyperparameter}
\end{figure*}

\paragraph{Triple Classification.}
We further evaluate GMT on triple classification over UMLS, CoDeX-S, and FB15K-237N, using a task-specific instruction template $\mathcal{I}_{GMT}$ that differs from link prediction (\autoref{prompt}). As shown in Table~\ref{tab:main_reasults}, GMT delivers consistent state-of-the-art performance across the three benchmarks. In particular, GMT achieves the best overall results on \textbf{UMLS} (Acc \textbf{94.55}, F1 \textbf{93.76}), ranks first on \textbf{CoDeX-S} (Acc \textbf{89.01}, F1 \textbf{87.61}), and also leads on the more challenging \textbf{FB15K-237N} (Acc \textbf{84.10}, F1 \textbf{87.02}). These results demonstrate that GMT generalizes beyond link prediction and remains robust across datasets with diverse relational patterns and label distributions.

\subsection{Ablation study: RQ2}

To assess the contribution of each component, we compare GMT with two variants: \textbf{w/o Semantics}, which replaces SGM with conventional pre-trained KGE embeddings, and \textbf{w/o Fusion}, which removes the Graph-as-Memory Cross-Attention Fusion Module and injects memory by simple prefix concatenation.

Table~\ref{tab:ablation} shows that both components are necessary. Removing SGM (\textbf{w/o Semantics}) consistently degrades performance on WN18RR and FB15k-237, indicating that relation-centric, context-aware semantics provide a stronger query-specific signal than static KG embeddings. More importantly, replacing cross-attention fusion with prefix injection (\textbf{w/o Fusion}) causes the largest drop (e.g., FB15k-237 MRR: 0.488 to 0.425), confirming that deep, token-wise memory retrieval is crucial and cannot be substituted by shallow concatenation.

\paragraph{Scoring Function in Stage-1 Pre-training.}
We further study the scoring function $F_{\mathcal{G}}(h,r,t)$ used to pre-train SGM. As reported in Table~\ref{tab:score}, a RotatE-style scoring function performs best on FB15k-237 (MRR \textbf{0.471}), while alternatives such as TransE/DistMult/ComplEx or an MLP lead to lower accuracy. We therefore adopt RotatE by default in our implementation.
The superiority of RotatE can likely be attributed to its inherent ability to model complex relational patterns.

\begin{table}[t]
\small
\centering
\caption{The scoring function for the link prediction task.}
\label{tab:score}
\setlength{\tabcolsep}{7pt}
\renewcommand{\arraystretch}{1.0}

\definecolor{lightgray}{gray}{0.95}
\begin{tabular}{c|cccc}
\hline
\textbf{Model} & \textbf{MRR} & \textbf{Hits@1} & \textbf{Hits@3} & \textbf{Hits@10} \\ \hline
\rowcolor{lightgray}
\multicolumn{5}{c}{\textbf{FB15k-237}} \\ \hline
GMT(w/ RotaE) & \textbf{0.471} & \textbf{0.380} & \textbf{0.488} & \textbf{0.623} \\
w TransE & 0.459 & 0.375 & 0.483 & 0.610 \\
w DistMult & 0.454 & 0.366 & 0.481 & 0.615 \\ 
w ComplexE & 0.455 & 0.368 & 0.481 & 0.613 \\
w MLP & 0.447 & 0.358 & 0.475 & 0.606 \\
\hline
\end{tabular}
\end{table}

\begin{table}[t]
\small
\centering
\setlength{\tabcolsep}{6pt}
\renewcommand{\arraystretch}{1.0}
\caption{Impact of LLM Relational Knowledge Enhancement.}
\label{tab:knowledge-enhancement}

\definecolor{lightgray}{gray}{0.95}
\begin{tabular}{c|cccc}
\hline
\textbf{Model} & \textbf{MRR} & \textbf{Hits@1} & \textbf{Hits@3} & \textbf{Hits@10} \\ \hline
\rowcolor{lightgray}
\multicolumn{5}{c}{\textbf{FB15k-237}} \\ \hline
GMT & \textbf{0.488} & \textbf{0.394} & \textbf{0.505} & \textbf{0.629} \\
w/o Enhancement & 0.454
& 0.370 & 0.475 & 0.601 \\
\hline
\end{tabular}
\end{table}

\begin{table*}[t]
\small
\centering
\renewcommand{\arraystretch}{0.9}
\setlength{\tabcolsep}{18pt}
\caption{
    Case study: comparison of the top-5 neighbor relations for a query, before and after applying Knowledge Enhancement. Abbreviations stand for full relation names, e.g., \textbf{Gov Position (Title)} for \texttt{.../government\_position\_held/basic\_title}. The Change column highlights the significant re-ranking based on semantic understanding.
}
\label{tab:case_study}
\begin{tabular}{c c c c c >{\columncolor{gray!20}}c@{}} %
\toprule
\multicolumn{2}{c}{\textbf{Without Knowledge Enhancement (Lexical Matching)}} & \multicolumn{3}{c}{\textbf{With Knowledge Enhancement (Semantic Relevance)}} \\
\cmidrule(r){1-2} \cmidrule(l){3-5}
\textbf{Neighbor Relation} & \textbf{Score} & \textbf{Neighbor Relation} & \textbf{Score} & \textbf{Change} \\
\midrule
Gov Position (Title)        & 0.859 & Gov Position (Title)        & 0.611 & \textbf{---} \\
Gov Position (Sessions)     & 0.828 & Gov Position (Sessions)     & 0.460 & \textbf{---} \\
Person (Profession)         & 0.250 & \textbf{Person (Nationality)} & 0.385 & \textcolor{blue}{\textbf{↑ Up (New)}} \\
Person (Places Lived)       & 0.206 & Person (Employment)         & 0.377 & \textcolor{blue}{\textbf{↑ Up}} \\
Person (Employment)         & 0.184 & \textbf{Person (Profession)}  & 0.338 & \textcolor{red}{\textbf{↓ Down}} \\
\bottomrule
\end{tabular}
\end{table*}

\paragraph{Hyperparameter Sensitivity.}
We analyze the sensitivity of GMT to three key hyperparameters in \autoref{fig:hyperparameter}: (i) Top-$K$ neighbor selection in SGM, (ii) the number of graph memory tokens $m$, and (iii) the number of top Transformer layers equipped with memory cross-attention. \textbf{Top-$K$} shows a clear sweet spot at a moderate neighborhood size (around $K{=}8$--$16$); too small loses context, while too large introduces noise and degrades MRR/Hits@1. 
Increasing the number of \textbf{memory tokens} improves performance and saturates around $m{=}32$, after which gains become marginal. 
Finally, injecting memory into more \textbf{top Transformer layers} consistently boosts all metrics, with the best results at the largest setting, supporting the benefit of deep, multi-layer memory retrieval.

\begin{table*}[h]
\footnotesize
    \centering
    \caption{Robustness analysis across different LLM generators. Best results are in \textbf{bold}, second best are \underline{underlined}.}
    \label{tab:llm_robustness}
    \renewcommand{\arraystretch}{1.0} 
    \setlength{\tabcolsep}{5pt} 
    {
        \begin{tabular}{c|l|cccc|cccc|cccc}
        \toprule
        \multirow{2}{*}{\textbf{Source}} & \multirow{2}{*}{\textbf{LLM Generator}} & \multicolumn{4}{c|}{\textbf{WN18RR}} & \multicolumn{4}{c|}{\textbf{FB15k-237}} & \multicolumn{4}{c}{\textbf{FB15k-237N}} \\
        \cmidrule(lr){3-6} \cmidrule(lr){7-10} \cmidrule(lr){11-14}
        & & \textbf{MRR} & \textbf{H@1} & \textbf{H@3} & \textbf{H@10} & \textbf{MRR} & \textbf{H@1} & \textbf{H@3} & \textbf{H@10} & \textbf{Acc} & \textbf{P} & \textbf{R} & \textbf{F1} \\
        \midrule
        \multirow{3}{*}{\textbf{Closed}} 
        & GPT-4o (Ours)     & \textbf{0.621} & \underline{0.569} & \textbf{0.667} & \textbf{0.703} & \underline{0.488} & \textbf{0.394} & \underline{0.505} & \underline{0.629} & \textbf{84.10} & \textbf{83.14} & \underline{91.27} & \textbf{87.02} \\
        & Claude-3.5-Sonnet & \underline{0.617} & \textbf{0.572} & \underline{0.660} & \underline{0.702} & \textbf{0.490} & \underline{0.390} & \textbf{0.511} & \textbf{0.635} & \underline{84.05} & \underline{83.11} & \textbf{91.35} & \underline{86.98} \\
        & Gemini-1.5-Pro    & 0.612 & 0.565 & 0.656 & 0.697 & 0.485 & 0.390 & 0.502 & 0.626 & 83.95 & 82.98 & 91.10 & 86.85 \\
        \midrule
        \multirow{3}{*}{\textbf{Open}} 
        & Qwen3-32B-Instruct  & 0.619 & 0.567 & 0.664 & 0.700 & 0.486 & 0.391 & 0.503 & 0.627 & 83.92 & 82.95 & 91.08 & 86.82 \\
        & Qwen3-8B-Instruct   & 0.613 & 0.561 & 0.658 & 0.694 & 0.481 & 0.386 & 0.497 & 0.620 & 83.52 & 82.55 & 90.68 & 86.42 \\
        & Llama-3-8B-Instruct & 0.610 & 0.558 & 0.655 & 0.690 & 0.478 & 0.383 & 0.494 & 0.616 & 83.25 & 82.26 & 90.35 & 86.11 \\
        \bottomrule
        \end{tabular}
    }
\end{table*}

\subsection{Analysis of Semantics: RQ3}

A key design in SGM is relation knowledge enhancement: we use an LLM to generate relation definitions and embed them as semantic vectors for Top-$K$ neighbor selection. We evaluate \textbf{GMT w/o Enhancement}, which encodes only raw relation names with the same Sentence-BERT.

\paragraph{GMT w/o Enhancement.} As shown in Table~\ref{tab:knowledge-enhancement}, removing enhancement consistently degrades performance on FB15k-237 (MRR drops from 0.488 to 0.454; Hits@1 drops from 0.394 to 0.370), verifying that relation names alone provide insufficient semantics for reliable neighbor filtering.

\paragraph{Robustness Across LLM Generators.} We further investigate whether GMT relies on specific proprietary models for definition generation. Table~\ref{tab:llm_robustness} compares performance across various closed-source and open-source LLMs. Results show that GMT maintains consistent performance, with even lightweight open-source models yielding negligible drops. This confirms that GMT benefits from the explicit semantic guidance rather than the capabilities of a specific model.

\paragraph{Case Study.}
To highlight the role of relational semantics, we examine the query (Barack Obama, /government/politician/government\_positions\_held..., ?), which describes the geographical or administrative scope of a political position. 
As shown in Table~\ref{tab:case_study}, knowledge enhancement substantially re-orders the top-5 neighbors, shifting retrieval from surface lexical matching to semantic relevance. Without enhancement, the ranking is dominated by relations with shared string prefixes (e.g., \texttt{Gov Position (Title)} and \texttt{Gov Position (Sessions)}), which captures naming overlap rather than the intended semantics.
With knowledge enhancement, neighbor retrieval becomes semantics-driven. In particular, \texttt{Person (Nationality)} appears and ranks highly, aligning with the query’s notion of jurisdiction rather than string overlap. Meanwhile, semantically related relations (e.g., \texttt{Person (Employment)}) are promoted and less relevant ones (e.g., \texttt{Person (Places Lived)}) are suppressed, yielding a cleaner evidence set. Overall, the case study shows that SGM goes beyond lexical matching and builds a more semantically coherent graph memory, providing more reliable signals for LLM reasoning.

\section{Conclusion}

In this paper, we propose Graph-as-Memory Tuning (GMT), a memory-centric framework that integrates knowledge graphs with LLMs beyond shallow prefix fusion. GMT builds query-specific graph memory tokens using a Semantic Graph Module (SGM) with knowledge-enhanced relation semantics, and injects them into multiple Transformer layers via a Graph-as-Memory Cross-Attention Fusion Module, enabling token-wise retrieval of graph evidence during generation. We further adopt LoRA on memory cross-attention for parameter-efficient adaptation with a frozen base LLM.

Experiments on link prediction and triple classification benchmarks show that GMT achieves state-of-the-art or highly competitive performance, and ablations confirm the effectiveness of both SGM and cross-attention fusion, while additional analyses show GMT is robust to different LLMs used for relation knowledge enhancement.
Future work will explore (i) richer graph memory construction to support deeper reasoning, and (ii) extending GMT to broader knowledge-intensive generation settings beyond KGC.







\section{AI Usage Statement}
We use LLMs only for writing refinement (readability and grammar). No LLM is involved in our code or experiments. All results are produced and verified by the authors.

\appendix
\section*{Appendix}
\section{Prompt Template}
\label{prompt}

\begin{promptbox}
{Prompt: Knowledge Enhancement}
\footnotesize
\textbf{Task:} Generate a canonical and descriptive definition for a knowledge graph relation by following the provided examples. The definition must clearly explain the semantic relationship in a neutral, factual sentence.
\vspace{1mm}
\hrule 
\vspace{1mm}
\textbf{Relation Name: }located\_in

\textbf{Definition:} The head entity is a smaller geographical, physical, or conceptual area that is situated within the larger area of the tail entity.
\vspace{1mm}
\hrule 
\vspace{1mm}
\textbf{Relation Name: }{relation\_name}

\textbf{Definition:} output
\end{promptbox}

\begin{promptbox}{Prompt: Instruction  Template for Experiments}
\footnotesize
\label{prompt:link}
\textbf{Link Prediction Instruction:} This is a knowledge graph completion task. Your goal is to predict the tail entity for an incomplete query triplet.
\vspace{1mm}
\hrule 
\vspace{1mm}
\textbf{Problem Definition:}
\begin{itemize}[noitemsep,topsep=0pt,parsep=0pt,partopsep=0pt,leftmargin=*]
    \item \textbf{Query:} ({head\_entity}, {relation}, ?)
    \item \textbf{Candidate Entities:} \{candidate\_1, ..., candidate\_20\}
\end{itemize}

\vspace{1mm}
\hrule 
\vspace{1mm}
\textbf{Response Specification:}
Analyze the provided candidates and return a ranked list of the top three most likely answers, from most to least probable.
\vspace{1mm}
\hrule 
\hrule 
\vspace{2mm}

\textbf{Triple Classification Instruction:} This is a triple classification task in knowledge graph. Your goal is to determine the correctness of the given triple.
\vspace{1mm}
\hrule 
\vspace{1mm}
\textbf{Input:} The triple is: (head\_entity, relation, tail\_entity)
\vspace{1mm}
\hrule 
\vspace{1mm}
\textbf{Output:} Please response True or False.
\end{promptbox}





\bibliographystyle{named}
\bibliography{ijcai2026}

\end{document}